\documentclass[conference]{IEEEtran}
\IEEEoverridecommandlockouts
\usepackage{cite}
\usepackage{amsmath,amssymb,amsfonts}
\usepackage{algorithmic}
\usepackage{graphicx}
\usepackage{textcomp}
\usepackage{xcolor}
\usepackage{multirow}
\def\BibTeX{{\rm B\kern-.05em{\sc i\kern-.025em b}\kern-.08em
    T\kern-.1667em\lower.7ex\hbox{E}\kern-.125emX}}
\begin{document}

\title{Research on Personalized Compression Algorithm for Pre-trained Models Based on Homomorphic Entropy Increase\\
}

\author{\IEEEauthorblockN{1\textsuperscript{st} Yicong Li}
\IEEEauthorblockA{\textit{Anhui University} \\
\textit{School of Computer Science} \\
and Technology \\
HeFei, China \\
summeryicong@163.com}
\and
\IEEEauthorblockN{2\textsuperscript{nd} Xing Guo}
\IEEEauthorblockA{\textit{Anhui University} \\
\textit{School of Computer Science} \\
and Technology \\
HeFei, China \\
guoxingahu@qq.com}
\and
\IEEEauthorblockN{3\textsuperscript{rd} Haohua Du}
\IEEEauthorblockA{\textit{Beihang University} \\
\textit{School of Computer Science} \\
and Technology \\
Beijing, China \\
duhaohua@buaa.edu.cn}
}

\maketitle

\begin{abstract}
In this article, we explore the challenges and evolution of two key technologies in the current field of AI: Vision Transformer model and Large Language Model (LLM). Vision Transformer captures global information by splitting images into small pieces and leveraging Transformer's multi-head attention mechanism, but its high reference count and compute overhead limit deployment on mobile devices. At the same time, the rapid development of LLM has revolutionized natural language processing, but it also faces huge deployment challenges. To address these issues, we investigate model pruning techniques, with a particular focus on how to reduce redundant parameters without losing accuracy to accommodate personalized data and resource-constrained environments. In this paper, a new layered pruning strategy is proposed to distinguish the personalized layer from the common layer by compressed sensing and random sampling, thus significantly reducing the model parameters. Our experimental results show that the introduced step buffering mechanism further improves the accuracy of the model after pruning, providing new directions and possibilities for the deployment of efficient and personalized AI models on mobile devices in the future.
\end{abstract}


\section{Introduction}
Vision Transformer (ViT) handles image classification tasks by splitting images into small pieces and embedding them linearly into sequences. The model utilizes Transformer's multi-head attention mechanism and self-attention mechanism to capture the global features of the image. While ViT performs well in terms of accuracy, its large number of parameters and computational overhead make it challenging to deploy on mobile devices. Therefore, model pruning technology has become the focus of recent research in order to reduce the computational requirements and parameters of models.

In addition to this, large language models (LLMS) have revolutionized the field of artificial intelligence, enabling natural language processing tasks previously thought to be unique to humans. Over the past few years, large language models (LLMS) have evolved from mere research artifacts [1] to useful products [2]. In large part, this evolution can be attributed to the dramatic increase in the scale of resources devoted to training [4]. The rapid evolution of large language models (LLMS) has led to architectures with billions to trillions of parameters, posing significant deployment challenges due to their large demands on memory, processing power, and energy consumption. Among them, the most excellent performance of the language large model is the Qwen language large model, including Qwen(basic pre-trained language model) and Qwen-chat (chat model fine-tuned using human alignment technology). Basic language models consistently show superior performance in numerous downstream tasks, while chat models, especially those trained using human feedback reinforcement learning (RLHF), are highly competitive. There is also a growing trend to prune models to meet the challenges posed by Vision Transformer and large language models (LLMS).

Since everyone's mobile device generates a lot of personalized data during its usage, it becomes increasingly attractive to store the data locally on the user/client/edge device and perform Machine Learning (ML) model training computations on the device with locally available data and occasional communication with aggregated parameter servers. This approach to training ML models is called Federated Learning\cite{r1}. One of the major challenges faced by Federated Learning is how to effectively replicate the global model locally at each client based on the personalized data.To address this challenge of Federated Learning, \cite{r2} had proposed to divide the deep learning model into base layers + personalization layers as a way to capture the personalization aspect of federated learning and personalize the deployment of the global model.

When dealing with data, the datasets owned by individuals cover a very small amount of data, typically hundreds or thousands of sheets, and the manual processing of markers is very time consuming, so the general operation is to train a model on a large dataset (e.g. cifar10, ImageNet), and then use the parameters trained by that model as the initial values of a model or a feature extractor, i.e., the pre-training parameters . Training a model based on this parameter saves time and computational resources, and also achieves faster training results.

The current pruning work of ViT model and LLM model pays more attention to how to delete redundant data without losing accuracy. However, in the actual deployment and application of ViT model and LLM model, the training parameters of the pre-trained model are trained based on the sum of various data sets in the network, although some redundant features are deleted. However, the training data parameters still contain redundant classification features, and the training parameters are huge. For some devices, such as mobile devices, the number of parameters is still too large, and different users have different habits and aesthetic styles. Personal mobile devices are more personalized than the data set of pre-training models. Therefore, when our ViT model and LLM model are deployed and applied, we need to pay more attention to personalized data with as few parameters as possible to improve the accuracy of personalized data. For this work, our goal is how to make the pre-training parameter model more targeted to personalized data by pruning the pre-training model, while subtracting unimportant parameters and maintaining the accuracy of the personalized model, which is the research goal discussed in this paper.

In this work, we propose a new pruning strategy called hierarchical pruning. This work prunes the model by studying the personalization layer and the generic layer of the model so that it achieves classification and discrimination of personalized data with the least number of parameters. In order to distinguish the personalized layer from the generic layer, we introduce a compressed perception approach, assuming that the personalized layer is sparse, we use random subsampling to randomly sample each linear layer, and obtain the personalized and generic layers based on the observations obtained from the random sampling. In order to achieve our desired pruning effect, we set different pruning rates according to the layers. In practice, we also discovered the step buffer mechanism and applied it to our pruning strategy, which greatly improved our accuracy after pruning.

This work goes to find the personalized layer and the generalized layer by compressed perception on a pre-trained model, aiming to provide a new idea and method for the next research work, and laying a foundation for the later research.

Our main contributions are as follows:

\begin{itemize}
    \item A new pruning algorithm is proposed, which can be used in both ViT and LLM models. Based on compressed sensing method, the personalized layer and the general layer are distinguished, and different degrees of pruning are carried out according to the importance of the personalized layer and the general layer.
    \item The degree and accuracy of pruning in ViT model and LLM model are systematically analyzed.
    \item We demonstrate that our proposed pruning method is well suited to pruning and personalization tasks: we demonstrate the reliability of our proposed method on different models and different data sets.
\end{itemize}

\section{Related Work}

\subsection{Vision Transformer Models}

Influenced by Transformer's success in the NLP field, more and more research is being done to apply it to visual tasks. CORDONNIER et al.\cite{r3} proposed a neural network structure that combines self-attention mechanisms with convolutional layers to overcome the limitations of convolutional layers. WU et al. \cite{r4} used semantic vision tokens instead of pixels and intensively modeled the relationships between tokens in Transformer, showing that visual Transformer outperforms CNN on some tasks. DOSOVITSKIY et al. \cite{r5} proposed a new vision Transformer architecture and demonstrated its high accuracy and good generalization on multiple datasets.

After the Vision transformer technology has been maturing, some recent researches have also made some variants of Vision transformer, including Touvron et al. who proposed a new Transformer model architecture, DeiT (Data-efficient image transformers), this model improves the performance of the student model by transferring the knowledge from a large Teacher model to a small Student model through knowledge distillation technique, and increases the diversity of training data by using data augmentation technique to perform mixup operation between different image blocks, which makes DeiT efficiency and accuracy\cite{r6}.Zhou et al. proposed an effective solution called Re-attention to solve the Attention Collapse problem when the model deepens the depth of the network,Re-attention can increase the diversity of Attention Maps, which allows the ViT model to deeper for training and obtain better performance\cite{r7}.

Personalized Learning Customizes learning plans according to students' needs and abilities. In 2018, Google introduced the concept of Personalized Federated Learning (PFL) to advance distributed machine learning and enhance privacy protection. Shamsian et al. \cite{r8} proposed the pFedHN approach, which allows each client to train the model using its own data distribution. Tan et al. \cite{r9} categorically discuss PFL methods and their future directions. Guangyu Sun et al. \cite{r10} discussed the personalization of vision Transformer (ViT) in federated learning, and introduced Prefix plug-in to personalize the self-attention layer of ViT, thereby improving the model performance. In this paper, personalization is introduced into the linear layer of ViT and the number of model parameters is reduced for easy deployment.

\subsection{Pruning for Transformers}

Despite the success of Transformer models in a variety of tasks, their high memory and computational resource requirements have hindered their implementation on resource-limited devices such as cell phones\cite{r11}. In order to improve model efficiency, many recent works have performed structural pruning and unstructural pruning on visual transformer models.Mukherjee et al. pruned the model through knowledge distillation techniques, which consisted of two main phases: 1) a pre-training phase, in which the authors pre-trained the model by using a large-scale base model (e.g. mBERT or XLM-R); and 2) a fine-tuning phase, in which the authors transfer the knowledge from the pre-trained model to a smaller target model through transfer learning and distillation techniques\cite{r12}. In this process the authors also used some techniques and strategies such as sample selection, teacher model bootstrapping, and asymmetric distillation to improve the performance of the model.Cheong et al. applied the k-means method derived by Han et al. and the authors' proposed binarization method based on Lam [23] to the Vision Transformer model and implemented iterative magnitude pruning, demonstrating the powerful compressibility of the Vision Transformer model\cite{r13}.

The researchers found that the Transformers model has a large number of redundant parameters, and removing these redundant parameters can save memory and computing resources. Michel et al.\cite{r14} found that while multiple attention heads were used for training, removing most of them for testing had little effect on performance. Tang et al. \cite{r15} proposed the patch slimming method to reduce the computation amount by identifying redundant patches. FAN et al. introduce a structured dropout method that is able to select subnetworks from large models without fine-tuning to improve efficiency\cite{r16}. Molchanov et al.\cite{r17} proposed an iterative pruning algorithm that optimizes the network through multiple rounds of pruning and fine-tuning. Our work is based on the iterative magnitude pruning technique to prune the model, but our work focuses on finding the personalized layer versus the generalized layer first and then performing iterative magnitude pruning on it.

\subsection{Pruning for LLMs}

In order to improve the number of parameters in language Large model (LLM), pruning technique has become a key strategy to optimize llm. Maintain model performance while reducing model size and computational costs. Gromov et al.\cite{r20} identify the best layer blocks to prune by considering cross-layer similarities, and make minor tweaks to the model in order to reduce the damage to the model, but the paper does not go into depth on how to make llm more efficient with the parameters in its deepest layers. Zhong L et al.\cite{r21} achieved fine-grained pruning by targeting redundancy in multi-head attention (MHA) and multi-layer perceptron (MLP) blocks, splitting each Transformer layer into MHA and MLP blocks, using confusion metrics to assess the importance of these blocks, and iteratively pruning the model based on the importance assessment.

The QWEN series of models, on the other hand, exhibit excellent performance due to their advanced tool usage and planning capabilities for creating proxy applications. This series includes a range of parameters from 500 to 72 billion, with intensive models and expert hybrid models. In order to explore the pruning effect of our strategy on large language models, we selected the QWEN series of large language models.

\section{Method}

\subsection{Problem definition}
For our work, we want to personalize the pre-trained model by pruning it for the personalized data of individualized users. So we need to collect personalized data first and then load the pre-trained model and prune the pre-trained model to different degrees by dividing it into personalized and generic layers. We define the personalized user data i.e. our input as $D_{\text {user }}$, and the output as the personalized model $M_{\text {user }}$.

\begin{equation}\label{...}
M_{\text {user }}=\mathrm{M}\left(D_{\text {user }}, \mathrm{W} \mid w^{\prime}=0\right)
\end{equation}

where W is our pre-trained model parameters, M is the pre-trained model, and $w^{\prime}$ is the redundant parameters that need to be pruned, as shown in Fig. 1.

\begin{figure}[htbp]
    \centering
    \includegraphics[width=4cm]{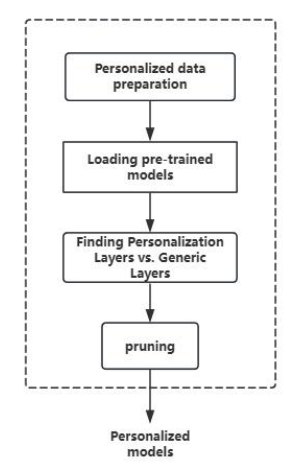}
    \caption{Model Pruning Flowchart}
    \label{fig:galaxy}
\end{figure}

\subsection{Vision Transformer}
Inspired by recent research, we found that for model pruning, the loss value has a simple and direct role for us to measure the importance of the weights, but since the model parameters are very large, the memory and computational overhead of measuring them one by one or in groups is also very large, so many variants are generated, such as Hansen's Matrix, Taylor Expansion, etc. related deformations. However, for our present work, we only study the linear layers in the model, as shown in Fig. 1, i.e., the mha and mlp parts of the model blocks. There are four linear layers included in a block, and assuming we have a total of L blocks, then we only need to study 4*L linear layers. For our work, we only need to find out the personalized layer and the generic layer among these 4*L linear layers.

\begin{figure*}[t]
\centering
\includegraphics[width=0.8\textwidth]{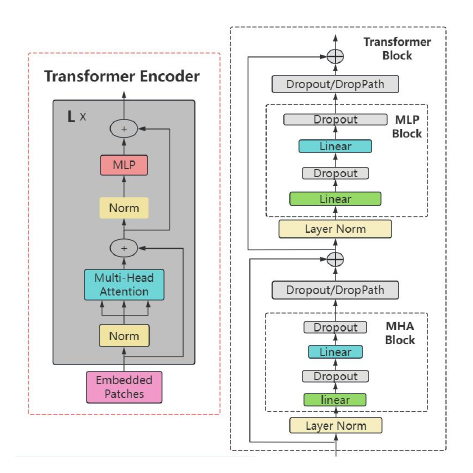} 
\caption{Vision Transformer Model.}
\label{fig2}
\end{figure*}

In order to analyze the importance of different linear layers uniformly so that we can define personalized and generic layers, we define a variable $l_{s}$ for each linear layer group s, where takes the value of 0 or 1, and statistically score the importance of each linear layer group by setting $l_{s}$ to 0. The weight $W_s$ of a linear layer group s is redefined as $W_s = l_{s}w_{s}$, which represents whether or not the weight of the linear layer group s is pruned.$L(D,l_{s}w_{s})$ represents the loss value of the linear layer group s on the dataset D, and the formula is shown below:

\begin{equation}\label{...}
L(D,l_{s}w_{s}) = L(D,W|w_{s} =0) - L(D,W) 
\end{equation}

where D represents the dataset and W represents the model weights. We consider the linear layer group as the personalized layer when the value of $L(D,l_{s}w_{s})$ is negative and the personalized layer group as the generic layer when its value is positive.

\subsection{Compressed Sensing}
Compressed perception is a method for efficiently obtaining sparse signals, the theory of compressed perception was first conceived by Candes a decade ago and with the help of mathematicians and DONOHO\cite{b18}, aims at reconstructing signals by sampling far less than the number of samples required by traditional methods. The idea of compressed perception is to exploit the sparsity or low-rank nature of signals, i.e., signals can be reconstructed at much lower than their Nyquist sampling rate, and thus the prerequisite for compressed perception needs to be satisfied that the signal is sparse. And we believe that in the migration of personalization task, the number of generic layers in a good model is much smaller than the personalization layer, so we take the generic layer as a sparse signal, and use the idea of compressed perception to divide the linear layers of the model, identify the personalization layer and the generic layer in the model, and then prune them.

In compressed sensing, if the sparse signal is sampled using random subsampling, it can be recovered using the algorithm of matched tracking. As shown in Fig. 3.

\begin{figure*}[t]
    \centering
    \includegraphics[width=4cm]{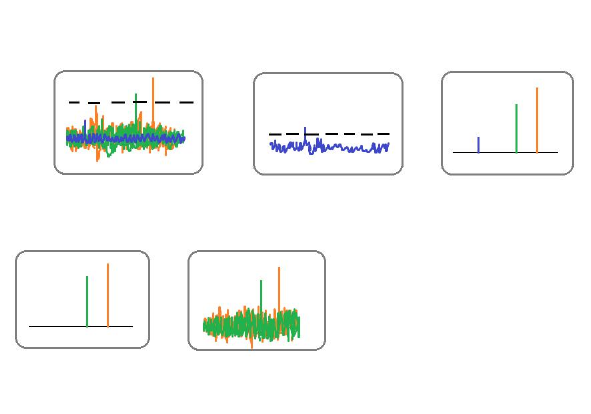}
    \caption{Compression-aware Signal Recovery}
    \label{fig:galaxy}
\end{figure*}

Assuming that e is the sparse signal and a is the noise signal obtained after our random sampling, we can detect the highest two signals first by setting two thresholds, and then use the noise data a to eliminate the noise generated by the two signals in b. The obtained, as shown in d, is the last kind of signal we need to detect. If the original sparse signal has more than three non-zero values, it can be solved one by one by iteration.

For this work, we believe that we can set the linear layer to 0 and get the loss value of the model by setting the stochastic matrix, where the loss value is our observation and the stochastic matrix is our observation matrix, and our goal is to solve for our generalized and personalized layers based on the known observation y and observation matrix $\varPhi$.

We solve for them by setting a threshold. According to the threshold value, our linear layer is divided as follows:

\begin{equation}\label{...}
layer=\left\{
\begin{array}{l}
Personalized \, Layer 
\\ \qquad\qquad\qquad\quad if \ loss_{layer_{s}=0 }  >threshold  
\\ Generic \, Layer  \quad  if \ loss_{layer_{s} = 0} < threshold \ 
\\ \qquad and \ layer_{s} \ not \ in \ Personalized \ Layer
\end{array}
\right.
\end{equation}

Where, the threshold threshold is the loss value when the model does not carry out any operation, the specific algorithm idea of personalized layer and generic layer division is: we through the random matrix way along with some of the linear layer all pruning, and then calculate the loss value of this model, according to the comparison of the loss value and threshold, we will divide the linear layer into personalized layer and generic layer. 
The implementation process of the layer classification algorithm is shown in algorithm 1.

\begin{figure}[t]
    \centering
    \includegraphics[width=9cm]{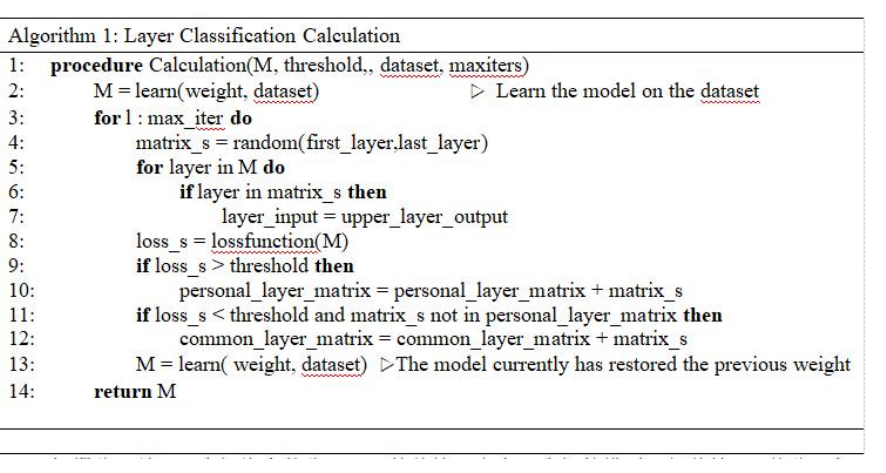}
    \label{fig:galaxy}
\end{figure}

Based on the layering results, we will assign different pruning ratios to them to prune our model as shown in the following equation, where prob is the custom pruning ratio and step\underline{~}prob is the layer pruning ratio.

\begin{equation}\label{...}
step\underline{~}prob=\left\{
\begin{array}{l}
prob  \qquad\qquad\qquad\quad\qquad Generic \; Layer  
\\ prob/2  \qquad\qquad\quad  Personalized \; Layers 
\\ \left ( prob+prob/2   \right ) /2  \qquad\qquad\quad\quad Others
\end{array}
\right.
\end{equation}

\section{Experimental Results}

\textbf{Dataset and Models} We conducted experiments on image classification on Cifar-10 which consists of 10 output classes, 50,000 training images and 10,000 test images\cite{b19}. On this image classification dataset, we have implemented a compression-aware model compression based approach using Vision Transformer, DeiT and Qwen model, we do not use any modes such as data augmentation, we do not use a distributed training approach, and we run only on a single gpu. For the LLM model experiment, we adopted the Qwen1.5-0.5B-Chat model in the QWEN series of models, which contains 24 Decoder layers and 50 million parameters. All experiments were performed on GeForce RTX 4090 and GeForce RTX 3080 Lite Hash Rate devices with batchsize of 32 and epoch set to 10.

\subsection{Experimental settings}

\textbf{Training Setup} The entire implementation of our method is divided into two main steps, taking the Vision Transformer model as an example.

(1)Step 1: Finding the personalized layer. We first use the random method to generate a random matrix, record the loss value of the model corresponding to this random matrix, and determine the personalized layer and the generic layer based on the comparison with the threshold value.

(2)Step 2: Pruning. Based on our personalized layer and generic layer classification results, we confirm the pruning ratio of each Linear layer in the ViT model and qwen model and perform iterative pruning.

\begin{table*}[t]
\begin{center}
\begin{tabular}{cccc}
\hline
\hline
\textbf{Model}&\textbf{Methods}& \textbf{Compression(\%)} & \textbf{Accuracy(\%)} \\
\hline
\multirow{3}{*}{ViT-B} & Base & - & 96.3 \\ 
& VTP & 44.0 & 90.61(-5.69) \\
& Ours & 44.8 & 92.9(-3.40) \\
\hline
\multirow{3}{*}{DeiT-B} & Base & - & 90.7 \\
& SViTE & 34.41 & 88.44 \\
& Ours & 22.80 & 89.0 \\
\hline
\multirow{3}{*}{Qwen1.5} & Base & - & 95.3 \\ 
& LaCo & 20.97 & 86.48(-8.82) \\
& Ours & 21.10 & 94.57(-0.73) \\
\hline
\hline
\end{tabular}
\caption{Pruning accuracy of different models and different methods}
\label{tab2}
\end{center}
\end{table*}

\textbf{Solution Implementation} We add the is\underline{~}skip variable as well as the layer\underline{~}number variable to each Linear layer in the ViT model and qwen model, we number the Linear layers in the ViT model by the layer\underline{~}number parameter, and the is\underline{~}skip parameter represents whether the current Linear needs to be pruned or not, and if it does, we prune all the current Linear layer is pruned. Use random method to generate a random matrix, if the layer\underline{~}number parameter of the Linear layer is in the random matrix, we set the is\underline{~}skip parameter to True, which represents that we prune all the current Linear layers, and calculate the loss value of the model after pruning. After the loop is executed n times, we divide all Linear layers into personalized and generic layers based on the threshold value. Then the pruning ratio is determined according to the division result, and the model is weighted and pruned. 

\textbf{Baselines} We selected three models, ViT-B, Deit-B and Qwen1.5, to verify the rationality and scalability of our scheme, and compared ourselves with previous compression methods, including VTP (Kumar et al.\cite{r19}, 2022), LaCo (Yang et al.\cite{r22}, 2024) and SViTE (Chen et al.\cite{r23}, 2021). 

\subsection{Main Results and Analysis}
Different results obtained by different pruning ratios for Vision Transformer model are shown in Table 2, and parameter optimization results are shown in Table 3. We conducted experiments on Prune Rate, Prune Object, and Random Number respectively, and the results of the experiments are shown in Table 3.The best results are obtained when the Random Number is 2, but the pruning ratio of the model is small.In order to weigh the balance between the pruning ratio and the accuracy, we defaulted to the Random Number in our experiments as 4. we notice that the compression based perceptual model compression method has a high pruning ratio of 44.8\% for the whole model when the Prune Rate is 0.05, but still maintains the accuracy at 92.9\%, which is enough to prove the robustness of our proposed method.

\begin{table*}[t]
\begin{center}
\scalebox{0.8}{
\begin{tabular}{|c|c|c|c|c|c|c|c|}
\hline
\textbf{Model}&\textbf{Prune Rate}&\textbf{Compression(\%)}&\textbf{Accuracy(\%)}&\textbf{Model}&\textbf{Prune Rate}&\textbf{Compression(\%)}&\textbf{Accuracy(\%)} \\
\hline
\multirow{6}{*}{Vision Transformer} & Baseline & - & 96.3 & \multirow{6}{*}{Qwen Model} & Baseline & - & 95.3\\
\cline{2-4} \cline{6-8}
& 0.01 & 11.1 & 94.7(-1.6) & & - & - & - \\
\cline{2-4} \cline{6-8}
& 0.02 & 22.1 & 94.5(-1.8) & & 0.02 & 9.04 & 95.38(+0.08) \\
\cline{2-4} \cline{6-8}
& 0.03 & 33.7 & 94.0(-2.3) & & - & - & - \\
\cline{2-4} \cline{6-8}
& 0.04 & 44.8 & 92.9(-3.4) & & 0.04 & 21.1 & 94.57(-0.73) \\
\cline{2-4} \cline{6-8}
& 0.05 & 55.5 & 88.0(-18.3) & & 0.06 & 31.71 & 89.64(-5.66) \\
\hline
\end{tabular}
}
\caption{Pruning Accuracy for Different Prune Rates in the Vision Transformer Model and QWEN Model}
\label{tab1}
\end{center}
\end{table*}

\begin{table}[t]
\begin{center}
\scalebox{0.8}{
\begin{tabular}{|c|c|c|c|c|}
\hline
\textbf{Prune}&\textbf{Prune}&\textbf{Random}& \multirow{2}{*}{\textbf{Compression(\%)}} & \multirow{2}{*}{\textbf{Accuracy(\%)}} \\
\textbf{Rate}&\textbf{Object}&\textbf{Number}& &  \\
\hline
\multirow{3}{*}{0.05} & \multirow{3}{*}{weight} & 8 & 46.4 & 92.0 \\
\cline{3-5} 
&  & 4 & 55.5 & 88.0 \\
\cline{3-5}
&  & 2 & 59.7 & 85.2 \\
\hline
\multirow{2}{*}{0.04} & weight & 4 & 44.8 & 92.9 \\
\cline{2-5}
& gradient & 4 & 44.6 & 79.6 \\
\hline
0.02 & weight & 4 & 22.1 & 94.5 \\
\hline
0.01 & weight & 4 & 11.1 & 94.7 \\
\hline
\end{tabular}
}
\caption{Hyperparameter Tuning}
\label{tab2}
\end{center}
\end{table}

\textbf{Random Number Analysis}  We have conducted relevant experiments for the cases of Random Number of 8, 4, and 2, respectively, while ensuring that other hyperparameters remain unchanged, and the experimental results are shown in Table 4. Under the same pruning ratio, when the random number is smaller, the final pruning of the model is less and the accuracy is higher. First of all, as the random number in the random matrix increases sequentially, the more random combinations of personalized and generalized layers, the wider the distribution of results included in the classification, i.e., the number of personalized layer lists and generalized layer lists is increasing, and the number of intermediate layers is decreasing, then the overall pruning ratio increases. This law is not obvious when the Prune Rate is small, but it is clearly visible as the Prune Rate keeps increasing. Secondly, our model compression method is implemented based on the idea of compression perception, and the premise of compression perception is that the signal is sparse, i.e., the number of personalized layers is less than the number of generalized layers, so that the intermediate layers are more divided into the generalized layers, i.e., the pruning ratio of the layers increases, and the overall pruning ratio of the model increases.

\begin{table}[htbp]
\begin{center}
\begin{tabular}{|c|c|c|c|}
\hline
\textbf{Prune} & \textbf{Random} & \multirow{2}{*}{\textbf{Compression(\%)}} & \multirow{2}{*}{\textbf{Accuracy(\%)}} \\
\textbf{Rate} & \textbf{Number} &  & \\
\hline
\multirow{3}{*}{0.04} & 8 & 38.2 & 93.7\\
\cline{2-4} 
& 4 & 44.8 & 92.9 \\
\cline{2-4}
& 2 & 47.7 & 92.3 \\
\hline
\multirow{3}{*}{0.02} & 8 & 18.8 & 94.4\\
\cline{2-4} 
& 4 & 22.1 & 94.5 \\
\cline{2-4}
& 2 & 23.9 & 94.5 \\
\hline
\multirow{3}{*}{0.01} & 8 & 9.6 & 94.1\\
\cline{2-4} 
& 4 & 11.1 & 94.7 \\
\cline{2-4}
& 2 & 12.0 & 94.5 \\
\hline
\end{tabular}
\caption{Impact of random number of hyperparameters in random matrices on models}
\label{tab3}
\end{center}
\end{table}

\textbf{Prune Object Analysis} In addition to analyzing a random number of hyperparameters in a random matrix, we also performed comparative experiments on pruned objects of the model. There are many current Vision Transformer model compression methods, including weight-based pruning and gradient-based pruning. Because our compression-aware idea is used for personalization layer segmentation, we consider the personalization layer to be crucial for our classification task and the generic layer to be least important for our classification task, so we have the largest pruning ratio for the generic layer and the smallest for the personalization layer, and the premise on which this method is founded is based on the importance analysis of the personalization layer. For model compression, gradient pruning focuses more on the trend of weight change immediately after the change of two images in the classification task, rather than the importance of the weights themselves, which is contrary to our principle; while the principle of weight-based pruning method is precisely based on the importance of the weights for the classification task, so we adopt weight pruning instead of gradient pruning, and the experimental results also verify our theory. The comparison results are shown in Table 5:

\begin{table}[t]
\begin{center}
\begin{tabular}{|c|c|c|c|}
\hline
\textbf{Prune} & \textbf{Random} & \multirow{2}{*}{\textbf{Compression(\%)}} & \multirow{2}{*}{\textbf{Accuracy(\%)}} \\
\textbf{Rate} & \textbf{Number} & &  \\
\hline
\multirow{2}{*}{0.01} & weight & 11.1 & 94.7\\
\cline{2-4} 
& gradient & 11.3 & 93.0 \\
\hline
\multirow{2}{*}{0.02} & weight & 22.1 & 94.5\\
\cline{2-4} 
& gradient & 22.4 & 89.6 \\
\hline
\multirow{2}{*}{0.04} & weight & 44.8 & 92.9\\
\cline{2-4} 
& gradient & 44.6 & 79.6 \\
\hline
\end{tabular}
\caption{Analysis of weighted pruning and gradient pruning results}
\label{tab4}
\end{center}
\end{table}

\textbf{Personalization Layer Without} pruning Our method is implemented based on the division of the personalization layer and the generalization layer, and we believe that the personalization layer is crucial to our image classification task, so can we directly not prune the personalization layer and only prune the other layers?

In order to solve this doubt, we conducted a comparison experiment on whether to prune the personalization layer or not, as shown in Table 6. As can be seen from Table 6, when the pruning ratio of the whole model is basically the same, pruning the personalized layer is better compared to not pruning the personalized layer, which confirms the reasonableness of our method. From the table, we can see that when the pruning ratio is similar or even less, pruning the personalized layer is more accurate than not pruning the personalized layer, but in the case of pruning the weights, the difference in accuracy is only less than 1\%, which means that retaining the personalized layer and increasing the pruning ratio of the generalized layer greatly will have a difference in the effect, but it will not significantly reduce our accuracy, and this also confirms that we are more effective in pruning the personalized layer than in not pruning the personalized layer. On the other hand, it also confirms that our discrimination of personalized layer is based on the basis. For the model that does not cut the personalized layer, the accuracy will be slightly lower than the model that prunes the personalized layer, the reason is based on the increase of the pruning amplitude of a certain layer, when the pruning amplitude is higher than a certain degree, it will inevitably cause a certain negative impact on our model, so for the phenomenon of reduced accuracy, it is reasonable.

\begin{table}[htbp]
\begin{center}
\scalebox{0.7}{
\begin{tabular}{|c|c|c|c|c|}
\hline
\textbf{Prune} & \textbf{Prune} & \textbf{Whether Prune } & \multirow{2}{*}{\textbf{Compression(\%)}}&\multirow{2}{*}{\textbf{Accuracy(\%)}} \\
\textbf{Object} & \textbf{Rate} &  \textbf{Personal Layer} & & \\
\hline
\multirow{3}{*}{weight} & 0.04 & True & 44.8 & 92.9 \\
\cline{2-5} 
& 0.06 & False & 41.9 & 92.0 \\
\cline{2-5}
& 0.065 & False & 45.3 & 91.5 \\
\hline
\multirow{3}{*}{gradient} & 0.04 & True & 44.6 & 79.6 \\
\cline{2-5} 
& 0.065 & False & 44.0 & 77.2 \\
\cline{2-5}
& 0.0655 & False & 44.5 & 75.8 \\
\hline
\end{tabular}
}
\caption{Personalized Layer Pruning Comparison}
\label{tab5}
\end{center}
\end{table}

\section{Conclusion}
In this paper, we propose a Vision Transformer model compression method based on the idea of compressed perception, which can maintain a good accuracy with a lot of pruning on the model. Experiments show that this model compression method can greatly reduce the number of parameters in the model, speed up model inference, focus more on personalized data, and help achieve personalized deployment of large models. Our future work will introduce a small neural network model to this method, which makes the sparse matrix solution in the compression perception idea concrete, and introduce it into the formula for model loss calculation, which can eventually realize the unsupervised personalized pruning method with the lowest computational cost.


\begin{thebibliography}{00}
\bibitem{r1} McMahan B, Moore E, Ramage D, et al. Communication-efficient learning of deep networks from decentralized data[C]//Artificial intelligence and statistics. PMLR, 2017: 1273-1282.
\bibitem{r2} Arivazhagan M G, Aggarwal V, Singh A K, et al. Federated learning with personalization layers[J]. arXiv preprint arXiv:1912.00818, 2019.
\bibitem{r3} Cordonnier J B, Loukas A, Jaggi M. On the relationship between self-attention and convolutional layers[J]. arXiv preprint arXiv:1911.03584, 2019.
\bibitem{r4} Wu B, Xu C, Dai X, et al. Visual transformers: Token-based image representation and processing for computer vision[J]. arXiv preprint arXiv:2006.03677, 2020.
\bibitem{r5} Dosovitskiy A, Beyer L, Kolesnikov A, et al. An image is worth 16x16 words: Transformers for image recognition at scale[J]. arXiv preprint arXiv:2010.11929, 2020.
\bibitem{r6} Touvron H, Cord M, Sablayrolles A, et al. Going deeper with image transformers[C]//Proceedings of the IEEE/CVF international conference on computer vision. 2021: 32-42.
\bibitem{r7} Zhou D, Kang B, Jin X, et al. Deepvit: Towards deeper vision transformer[J]. arXiv preprint arXiv:2103.11886, 2021.
\bibitem{r8} Shamsian A, Navon A, Fetaya E, et al. Personalized federated learning using hypernetworks[C]//International Conference on Machine Learning. PMLR, 2021: 9489-9502.
\bibitem{r9} Tan A Z, Yu H, Cui L, et al. Towards personalized federated learning[J]. IEEE Transactions on Neural Networks and Learning Systems, 2022.
\bibitem{r10} Sun G, Mendieta M, Luo J, et al. Fedperfix: Towards partial model personalization of vision transformers in federated learning[C]//Proceedings of the IEEE/CVF International Conference on Computer Vision. 2023: 4988-4998.
\bibitem{r11} Han K, Wang Y, Chen H, et al. A survey on vision transformer[J]. IEEE transactions on pattern analysis and machine intelligence, 2022, 45(1): 87-110.
\bibitem{r12} Mukherjee S, Awadallah A. XtremeDistil: Multi-stage distillation for massive multilingual models[J]. arXiv preprint arXiv:2004.05686, 2020.
\bibitem{r13} Cheong R, Daniel R. transformers. zip: Compressing transformers with pruning and quantization[J]. Technical report, tech. rep., Stanford University, Stanford, California, 2019.
\bibitem{r14} Michel P, Levy O, Neubig G. Are sixteen heads really better than one?[J]. Advances in neural information processing systems, 2019, 32.
\bibitem{r15} Tang Y, Han K, Wang Y, et al. Patch slimming for efficient vision transformers[C]//Proceedings of the IEEE/CVF Conference on Computer Vision and Pattern Recognition. 2022: 12165-12174.
\bibitem{r16} Fan A, Grave E, Joulin A. Reducing transformer depth on demand with structured dropout[J]. arXiv preprint arXiv:1909.11556, 2019.
\bibitem{r17} Molchanov P, Tyree S, Karras T, et al. Pruning convolutional neural networks for resource efficient inference[J]. arXiv preprint arXiv:1611.06440, 2016.
\bibitem{r18} Donoho D L. Compressed sensing[J]. IEEE Transactions on information theory, 2006, 52(4): 1289-1306.
\bibitem{r19} Kumar A. Vision transformer compression with structured pruning and low rank approximation[J]. arxiv preprint arxiv:2203.13444, 2022.
\bibitem{r20} Gromov A, Tirumala K, Shapourian H, et al. The unreasonable ineffectiveness of the deeper layers[J]. arXiv preprint arXiv:2403.17887, 2024.
\bibitem{r21} Zhong L, Wan F, Chen R, et al. BlockPruner: Fine-grained Pruning for Large Language Models[J]. arXiv preprint arXiv:2406.10594, 2024.
\bibitem{r22} Yang Y, Cao Z, Zhao H. Laco: Large language model pruning via layer collapse[J]. arxiv preprint arxiv:2402.11187, 2024.
\bibitem{r23} Chen T, Cheng Y, Gan Z, et al. Chasing sparsity in vision transformers: An end-to-end exploration[J]. Advances in Neural Information Processing Systems, 2021, 34: 19974-19988.
\end{thebibliography}
\end{document}